\documentclass[a4paper, 12pt]{article}
\usepackage[english]{babel}
\usepackage[affil-it]{authblk}

\usepackage[letterpaper,top=2cm,bottom=2cm,left=2.5cm,right=2.5cm,marginparwidth=1.75cm]{geometry}

\usepackage{amsmath}
\usepackage{amssymb}
\usepackage{cite}  
\usepackage{graphicx}
\usepackage[usenames,dvipsnames]{color}
\usepackage[colorlinks=true, allcolors=blue]{hyperref}
\usepackage[justification=centering, font=footnotesize]{caption}
\usepackage{booktabs}
\usepackage{multirow}
\usepackage{listings}
\usepackage{array}
\newcolumntype{M}[1]{>{$\displaystyle\qquad}p{#1}<{$}}

\definecolor{comment}{rgb}{0, 0.6, 0}
\definecolor{keyword}{rgb}{0.96, 0.52, 0.02}
\definecolor{string}{rgb}{0.12, 0.4, 0.87}
\definecolor{background}{rgb}{0.95,0.95,0.92}
\lstdefinestyle{mypython}{
	language=Python,
    backgroundcolor=\color{background},   
    commentstyle=\color{comment},
    keywordstyle=\color{keyword},
    numberstyle=\tiny\color{magenta},
    stringstyle=\color{string},
    basicstyle=\ttfamily\footnotesize,
    breakatwhitespace=false,         
    breaklines=true,                 
    captionpos=b,
    keepspaces=true,
    numbers=left,
    numbersep=5pt,
	showspaces=false,
	showstringspaces=false,
    showtabs=false,
    tabsize=2
}
\usepackage{academicons}
\usepackage{xcolor}
\usepackage{orcidlink}

\newcommand{\myfig}[1]{{\color{black}Figure~\ref{#1}}}
\newcommand{\myeq}[2][e]{{\color{black}#1quation~(\ref{#2})}}
\newcommand{\mytab}[1]{{\color{black}Table~\ref{#1}}}
\newcommand{\mycite}[1]{{\color{black}\cite{#1}}}
\newcommand{\mycode}[1]{{\ttfamily\footnotesize{#1}}}

\newcommand{\mysec}[1]{{\color{black}section~\ref{#1}}}
\newcommand{\auth}[1]{{\normalsize#1}}

\newcommand{\tagsection}[1]{\section{#1}\label{#1}}
\newcommand{\tagsubsection}[1]{\subsection{#1}\label{#1}}

\title{Mathematical Computation on High-dimensional Data via Array Programming and Parallel Acceleration}

\author{
  \auth{Chen Zhang\orcidlink{0009-0007-7689-5030}}
  \thanks{E-Mail: \texttt{chen.zhang\_06sept@foxmail.com}}
}
\affil{OSF Project: \href{https://osf.io/dcs2h}{\textit{Open Science Informatics}}}

\begin{document}
\maketitle

\begin{abstract}
The development of hardware performance and the maturation of algorithm application have demonstrated that deep learning (DL) based frameworks offer significant advantages in the context of natural image and natural language processing tasks. However, there are significant challenges when attempting to replicate this success in other domains. In fact, the DL framework and its associated computation units are primarily designed for processing vectors and matrices, rather than data structures with higher dimension. The dimensionality curse, which arises when dealing with high-dimensional data, will result in heavy computation that is beyond the affordability of present computing systems. Despite some techniques proposed for supporting large-scale data, the majority of their corresponding tool chains are primarily designed for fulfilling descriptive statistics on data for business purposes, rather than for mathematical statistics which serve for downstream analysis.\par


To address this issue and fully leverage the modern techniques of parallel computation, we have proposed and implemented an innovative architecture for mathematical statistics.  This foundation is theoretically derived from the concept of space completeness.  The system supports the decomposition of high-dimensional data into dimension-independent intermediate structures and the subsequent allocation, processing, and final aggregation of computing tasks on the basis of different physical units or a computation cluster.   The implementation and associated infrastructures facilitate the seamless integration of a multitude of data mining and machine learning approaches into new workflows that are optimized by parallel acceleration.  The processing of this paradigm is not constrained by the structural properties of data, which makes it feasible to implement scientific computations on a majority of high-dimensional data types, such as natural or medical images, within an integrated framework with unification.\par

\hspace*{\fill}

\end{abstract}

\tagsection{Introduction}


The exponential growth of digitalization has led to the development of new techniques, in both domains of hardware (e.g., device miniaturization and semiconductor process) and software (e.g., data structure designed for efficient data query)\mycite{legner2017digitalization, agostini2020digitalization, frazier1995miniaturization, yugma2015integration, mao2003document, sedgewick2011algorithms}. Since the data governance becomes more standardized, with the decreasing cost of data acquisition as well as the increasing maturity of associated algorithms, the validation of approaches in the experimental stage that utilize data in a production environment becomes theoretically achievable. While the majority of manipulations and calculations on data in the production environment are undergoing the stage of applying essential statistical approaches on extremely large-scale data, which catalyzes the formation of some big data related techniques such as distributed computation\mycite{mascaro2021towards, siegel1985interconnection}.\par

Despite the substantial advantages of distributed frames as an representative solution for parallel acceleration, the current observations indicate that they are more suitable for simple data storage, retrieval, and modification in business contexts. However, they are not well-suited for mathematical computations in exploratory purposes\mycite{bertsekas2015parallel, kshemkalyani2011distributed}. The primary reason for this discrepancy is that the levels of abstraction between the target object to be processed through the database framework and the one in mathematical computation are significantly different. In database contexts, data frames are typically organized in a column-wise manner for fields, representing for distinct attributes. While the concept of observation refers for a certain row whose values can be filled in all the existing fields. The vertically stacked observations constitute the data frame in general sense. And the values inside a data frame can be the concrete numeric, the iterable object with certain length, or even pointers to more complicated object such as raw digital data\mycite{shanahan2015large, jacobs2016apache, saxena2017practical}. The distributed system has inherent advantage in the computation in terms of the data frame with large observations. Nevertheless there is the difference that the data manipulation in the context of mathematical computation refers specifically to the operations and calculations on numerical values, sets, or arrays.\par

The parallel computational acceleration based on distributed framework can be conceptualized as the implementation for computations with high efficiency on the matrix-like data frame, under the circumstance of the absolutely large number of samples\mycite{thoke2014theory, chow2006survey}. The majority of simple statistical tasks on database fields can benefit from the its cluster architecture. However, in terms of mathematical computation, it is in demand of efficient data mapping and more flexible intermediate structures. In most instances, the concept \emph{data} in mathematical computing is merely an numerical array stored in specific field of the data frame, which means the cluster and associated tool chains are not universal as for applying or improving the mathematical computations. Besides, the parallel computation in general-purpose is also of urgent need and profound significance, given the increasing resolution and dimensionality of data that should be processed.\par


This paper presents a novel programming paradigm and associated tool set designed to address the aforementioned issues in scientific computation. To ensure generalization and logical rigor, the tensor and Hilbert space were introduced as concepts in mathematics, serving as the infrastructural foundation for further coding implementations. To facilitate the execution of data mapping, we identified a pivotal intermediate, the melt matrix, as a row decoupled matrix whose structure can be disassembled and allocated to distinct parallel computing nodes. The repeated utilisation of the decoupling and coupling via this intermediate structure can efficiently reduce the effort of scientific computation on high-dimensional data, optimise mathematical computation, or even support further graphics processing unit (GPU) acceleration of the underlying operations.\par

\tagsection{Problem Definition and Solution}

A well-managed mathematical computing system should not merely satisfy the essential requirements of mathematical computation; it should also ensure the self-consistency of mathematical principles, facilitate the data processing with uncertain structures, and the maintainability of code and modules. Based on the coding practices observed in scientific computation, we have summarized the most common engineering challenges encountered in a range of aspects, from systematic design to technology selection progressively. The associated surveys, analyses, etc., used to solve specific challenges are also provided.\par

\tagsubsection{Infrastructural tech stack}

A comparison of mainstream, freely accessible programming languages, with a particular focus on the prototypical implementations for data analytics and further modeling, can be beneficial in determining the underlying architecture. The investigation results are presented in \mytab{tag1}. Values of items with asterisk(s) can be interpreted as the degree indicator as represented for that item. Check mark denotes a feature with sufficient implementation, otherwise that is under development or in lack of full support.\par

\begin{table}[htbp]
  \begin{center}
  \caption{Programming languages suitable for mathematics and statistics}
  \begin{tabular}{c|c c c c c c}
  \toprule
   - & \mycode{\textbf{Python}} & \mycode{\textbf{R}} & \mycode{\textbf{C}} & \mycode{\textbf{C++}} & \mycode{\textbf{Julia}} & \mycode{\textbf{Octave}} \\
  \midrule
  popularity & 1\ $\uparrow$ & 24\ $\downarrow$ & 2 & 4 & 29 $\downarrow$ & 50\mycode{+} \\ 
  \hline
  community & \mycode{*****} & \mycode{****} & \mycode{*****} & \mycode{*****} & \mycode{***} & \mycode{**} \\ 
  \hline
  numerical cal. & \mycode{numpy} & \mycode{built-in} & \mycode{gsl} & \mycode{gsl} & \mycode{GSL} & \mycode{built-in} \\
  \hline
  symbol cal. & \mycode{sympy} & - & - & - & \mycode{Symbolics} & \mycode{symbolic} \\ 
  \hline
  statistical cal. & \mycode{scipy} & \mycode{built-in} & - & - & \mycode{Gen} & \mycode{statistics} \\ 
  \hline
  functional & \checkmark & \checkmark & - & \mycode{C++17+} & \checkmark & - \\ 
  \hline
  linear algebra & \mycode{numpy} & \mycode{built-in} & \mycode{eigen} & \mycode{eigen} & \mycode{LinearAlgebra} & \mycode{built-in} \\ 
  \hline
  database frame & \mycode{pandas} & \mycode{built-in} & - & - & \mycode{DataFrame} & \mycode{built-in} \\ 
  \hline
  interdisciplinarity & \mycode{*****} & \mycode{****} & \mycode{**} & \mycode{***} & \mycode{***} & \mycode{****} \\ 
  \bottomrule
  \end{tabular}
  \label{tag1}
  \end{center}
\end{table}

The evaluation of the popularity and tendency of languages in \mytab{tag1} is based on the TIOBE index over the past decade\mycite{TIOBE2024}. Community vibrancy reflects the capability of the programming language's developer and operation team, which can pre-evaluate the potential difficulties in engineering realization. Numerical, symbolical, and statistical calculations for generic purposes are features of advanced mathematical computing systems. The corresponding implementation library is therefore listed for each programming language. The functional programming is a sophisticated paradigm that enables high scalability of programs. The extensive use of high-order functions allows the system to be extensible, supporting even unknown data in uncertain devices further, in a manner of comparatively lightweight modification on original code. Furthermore, the associated linear algebra system, the data frame structure for data analytics, and the system ecology of third-party packages in supporting different disciplinary usage are also under consideration.\par

Python is the preferred programming language for data analytics, machine learning (ML) and artificial intelligence (AI) due to its comprehensive set of advanced mathematical features and its suitability for multidisciplinary applications\mycite{harris2020array, ciaburro2019python, vasilev2019advanced, singh2020artificial}. The upper facilities of Python also exhibit a favorable trade-off at the data abstraction level and underlying implementation\mycite{perez2010python}.  For instance, the data frame structure proposed by the \mycode{pandas} development team has proven to be highly successful in various parallel or distributed architectures, including the \mycode{modin}, \mycode{dash}, and \mycode{spark}. In contrast, \mycode{C} and \mycode{C++} have accumulated significant technical debt. The limitation of their linear algebra system is a consequence of the strong assumption that the data is one- or two-dimensional. \mycode{R} and \mycode{Octave} excel in numerical computation, particularly in statistical calculations and linear algebra, respectively. Their data structures are primarily built-in, and both languages are part of the GNU project\mycite{stallman1998gnu, giorgi2022r, eaton1997gnu}. However, their comparatively closed architectures may present challenges for further integration.\par

\tagsubsection{Hilbert completeness and generalization}


Our problem statement, namely the undetermined shape of data and associated manipulations, can be further conceptualized under the analytical system in the context of abstract algebra. This system essentially interprets the dilemmas of extensibility of certain computation architectures. For some early algebraic systems, their designs are based on a limited-dimensional normed linear space rather than the Hilbert normed linear space. It is therefore not strictly closed with regard to some required operations, such like expansion for or aggregation from a dimension, in scientific computation. Both aforementioned spaces are of completeness that the concepts such as the inner product is natively included, with the bounded and unbounded property in terms of dimensionality, respectively. To be more precise, the feature that to be closed to the undetermined dimensionality including even infinite one, is the inherent attribute of the Hilbert space. For a computing system that is designed to support high-dimensional data, it is necessary that its computational principles should be reinterpreted on the basis of the Hilbert space as a generic form, in purpose of achieving that dimensionality closeness. And their application programming interfaces (APIs), should be designed correspondingly as generic functions in consideration of that generality.\par

These ideas are of profound significance with regards to underlying architecture design and implementation. The utilization of generic functions with scalability in dimensional variation enables the processing of data in diverse dimensions under an unified framework, regardless of modification to the basic architectures. It is exactly the most profitable thing of the Hilbert completeness based implementation, in the engineering practice for mathematical computation. Whereas this is not easy to be realized. The majority of the challenges now come from the generalization for computational principles. For instance, the $n$-dimensional sphere must be introduced, to accordantly integrate the concepts such like the line segment, the disc, the conventional sphere, or other structures in higher dimensions with property of rotation invariance. Another illustrative example can be found in \mytab{tag2}, where the Gaussian and its associated calculations defined in Hilbert space, should be implemented in more generalized form, i.e. multivariate Gaussian. The univariate or bivariate Gaussian distribution in this context, are nothing more than specific degenerated forms from the multivariate one.\par

\begin{table}[htbp]
  \begin{center}
  \caption{Generalization for Gaussian distribution and associated gradient}
  \renewcommand{\arraystretch}{2} 
    \begin{tabular}{c|c|c}
    \toprule
    - & \textbf{univariate} $\mathcal{N}(x|\mu, \sigma)$ & \textbf{multivariate} $\mathcal{N}(\boldsymbol{x}|\boldsymbol{\mu}, \boldsymbol{\Sigma})$ \\
    \midrule
    $p$ & $\displaystyle \frac{1}{\sqrt{2\pi}\sigma} \exp[-\frac{(x-\mu)^2}{2\sigma^2}]$ & $\displaystyle \frac{1}{(2\pi)^{\frac{k}{2}}|\boldsymbol{\Sigma}|^\frac{1}{2}} \exp [-\frac{1}{2}(\boldsymbol{x}-\boldsymbol{\mu})^\top \boldsymbol{\Sigma}^{-1} (\boldsymbol{x}-\boldsymbol{\mu})]$  \\
    \hline
    $\displaystyle \frac{\partial p}{\partial (x|\boldsymbol{x})}$ & $\displaystyle -\frac{\sigma^{-2}(x-\mu)}{\sqrt{2\pi}\sigma} \exp[-\frac{(x-\mu)^2}{2\sigma^2}]$ & $\displaystyle -\frac{\boldsymbol{\Sigma}^{-1} (\boldsymbol{x}-\boldsymbol{\mu})}{(2\pi)^{\frac{k}{2}}|\boldsymbol{\Sigma}|^\frac{1}{2}} \exp [-\frac{1}{2}(\boldsymbol{x}-\boldsymbol{\mu})^\top \boldsymbol{\Sigma}^{-1} (\boldsymbol{x}-\boldsymbol{\mu})]$ \\
    \bottomrule
    \end{tabular}
  \label{tag2}
  \end{center}
\end{table}

The generalization for computing principles results in the majority of computation steps being exclusively implemented in a relatively sophisticated manner, in which the algebraic concepts and associated calculations are massively introduced and applied (e.g., vectorization, matrix analysis and operation). This kind of paradigm will bring the \emph{buckets effect} to a system. Consequently, the dependencies on constructions with limited generalization may potentially become the Achilles' heel of a computation system designed for high-dimensional data.\par

\tagsubsection{Generic container}

From a mathematical point of view, the tensor is an ideal system for dealing with high-dimensional data structures. Booming researches studied on the related domains such as tensor analysis, underpins the theoretical supports in return, for its further applications. Most discussions about tensor can also satisfy the completeness aforementioned, since that the tensor and its associated operations can be considered as the basic data structure and corresponding universal algebra manipulation in accordance with the Hilbert space. For example, the singular value decomposition on matrix, is actually the  CANDECOMP/PARAFAC (CP) decomposition of the tensor applied to 2-dimensional situation\mycite{bergqvist2010higher, goulart2015tensor, cuppen1983singular}.\par

In terms of engineering fulfillment, Python provides the tensor-like data structure and manipulations for computation, through numpy. Since its first release from nearly 20 years before, array programming has become the de facto design criterion in the Python ecosystem, from basic algorithms to domain-specific utilities built on top\mycite{harris2020array, garrido2015introduction}. By affording features such as efficient data storage and access, easy broadcasting, syntax sugar for algebraic computation and etc., numpy is highly abstract, thus highly productive in implementation for developing generic functions.\par

Using the dense numpy array as the fundamental data-driven container, can not only ensure the self-consistent properties of the algebraic system, such as completeness, but also the extensibility when using frontier techniques. For example, sparse arrays utilized for natural language processing or topic modeling, dense matrices employed for computer vision, and sparse tensors utilized for user behavior analysis can be exchanged in a uniform manner via the numpy array. In fact, the choice of numpy and the use of array programming, can take full advantage in the Python ecosystem, since it is virtually the optimal solution for the maximum intersection among multidisciplinary domains, or multi-technical approaches.\par

\tagsubsection{Computational separability}

In order to gain insight into the successful applications of distributed systems in practice, it is essential to examine the rationale behind and the manner in which distributed computation can be applied. The Dynamic Data Frame (DDF), which is used as a basic data structure in most distributed systems, also corresponds to the design matrix in terms of data science\mycite{ponce2021ddf, perera2022high, sarkar2022parallelized}. Abstractly, a design matrix $\boldsymbol{M} \in \mathbb{R}^{n \times m}$ theoretically possesses the equivalent information as that of the family of its column-permutated matrices. A distributed system should be considered, under the condition that observation number of data (the scale of $n$ samples) is much over the affordability of a single node, in terms of storage as well as computation.\par

For any columnar partition $\boldsymbol{P} = \{\boldsymbol{P}_1,\ \dots,\ \boldsymbol{P}_s\}$ of the matrix $\boldsymbol{M}$, it at least satisfies the following three conditions:\par

\begin{itemize}
  \item $\boldsymbol{P}_i \in \mathbb{R}^{{k_i} \times m}$, where $n = \sum_{i=1}^s k_i,\ k_i > 0,\ \forall i \in \{1, \dots, s\}$
  \item $\forall i, j \in \{1, \dots, s\},\ i \neq j,\ \boldsymbol{P}_i \cap \boldsymbol{P}_j = \varnothing$
  \item $(\exists \boldsymbol{A} \in \{\mathrm{r}(\boldsymbol{A}) = n\}),\ \boldsymbol{A}([\boldsymbol{P}_1^\top\ \boldsymbol{P}_2^\top\ \cdots\ \boldsymbol{P}_{s-1}^\top\ \boldsymbol{P}_s^\top]^\top) = \boldsymbol{M}$
\end{itemize}

Where $[\boldsymbol{P}_1^\top\ \boldsymbol{P}_2^\top\ \cdots\ \boldsymbol{P}_{s-1}^\top\ \boldsymbol{P}_s^\top]^\top$ represents the vertical stack matrix of the partition $\boldsymbol{P}$. Since $\boldsymbol{P}$ is nothing but the set of observation subset from $\boldsymbol{M}$, the first two conditions establish that the $\boldsymbol{P}$ is a valid partition based on columns. And the last condition ensures the existence of an invertible matrix $\boldsymbol{A}$ that can somehow transform $\boldsymbol{P}$ into $\boldsymbol{M}$. With a proper partition, the computation of each part $\boldsymbol{P}_i$ in $\boldsymbol{P}$ can be physically affordable on its assigned node, through which manner a cluster can therefore facilitate the essential computation on data with a large scale.\par


In addition to the partitioning to reduce computation load, another interpretation of rationality is comes from the computational independence of approaches. The majority of algorithms that have been demonstrated on distributed systems make use of aggregation functions or combinations of these functions, which can be operated directly on both populations and samples. Some of those aggregation functions will not result in bias, with applying from samples towards population (for example, the maximum function), which are therefore perfectly suitable for use within the MapReduce framework\mycite{perera2022high}. Nevertheless, there are still part of operations that are involved in some sample-determined factors (e.g., the median, or the non-parameteric methods that relates to rank orders of samples). However for those operations, as the number of observations increases, the application of modern techniques such as randomization to some extent ensures that the statistical results derived from samples converge towards that of the population.\par

With regard to the tensor data structure, it is regrettable that there is no intermediate structure analogous to DDF that possesses both the partitionability of the original data and the computational independence among partitions, used for facilitating mathematical computation via parallel acceleration. Following an investigation of the majority of numerical computating steps on tensors, we have proposed an intermediary structure which is capable of satisfying all aforementioned constraints. This structure is suitable for fulfilling array programming on tensors, whether in a single computing node or a distributed cluster.\par

\tagsection{Framework and Practical Applications}

In order to ascertain the paradigmatical design of a computing system that allows the parallel operations upon independent physical units, a deconstruction of the execution steps of common numerical computations was performed. This understanding aimed to identify the generic form of intermediary structure that would facilitate the scientific computation. The \emph{melt matrix}, an intermediate container inspired by the behavior of a built-in function in the R language, is proposed as the readiness for mathematical computations via array programming, as well as the solution for tensor partition. The subsequent subsections will present its framework and design principles, followed by a series of illustrative implementation examples based on this architecture. All related concepts and demonstrations will primarily, but not limited to, be implemented in the contexts of the Python language.\par

\tagsubsection{Infrastructural architecture}

\begin{figure}[htbp]
  \centering
  \includegraphics[scale=1.05]{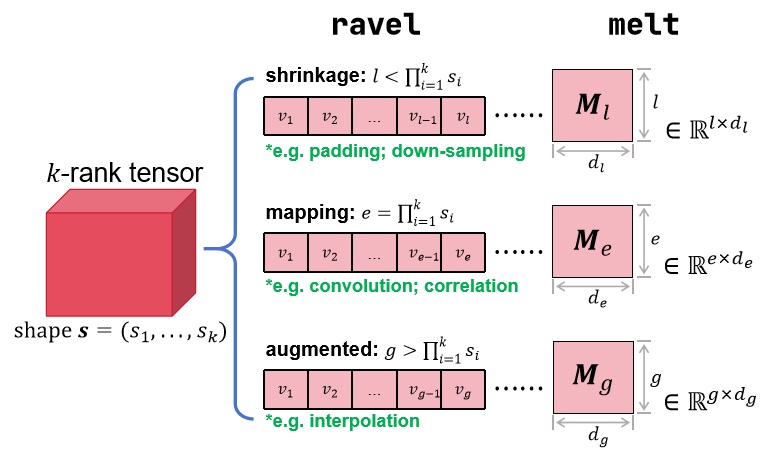}
  \caption{Structures after using ravel and melt operations respectively on a tensor}
  \label{tag3}
\end{figure}

Ravel is a type of operation that is used to flatten an iterable container into a numerical vector. In the context of scientific computation, the number of elements within a vector is not required to be completely equal to that of the unravel container. With the exception of global filtering, the application of techniques such as up- and down-sampling, or the use of a padding layer in DL architecture, all result in a reduction or expansion of the number of elements, as illustrated in \myfig{tag3}. In a similar manner, the melt matrix of a tensor is a two-dimensional array, wherein all numbers within its ravel vector are substituted by vectors with a fixed length ($d_l$, $d_e$, or $d_g$ in \myfig{tag3}). With regard to each row vector $\boldsymbol{M}_{i,:}$ in a melt matrix $\boldsymbol{M}$, it should be noted that the value included is not solely that corresponding to the element $v_i$ in its ravel vector $\boldsymbol{v}$. It also encompasses the values within the neighbourhood surrounding that $v_i$ inside the original tensor.\par

\begin{figure}[htbp]
  \centering
  \includegraphics[scale=1.1]{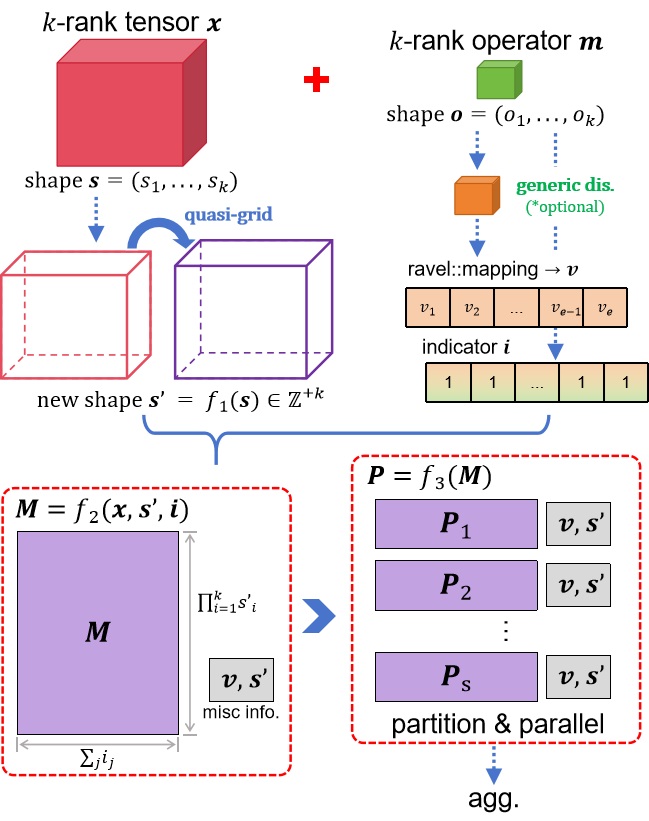}
  \caption{Schematic diagram of framework for parallel array programming}
  \label{tag4}
\end{figure}

The principal schematic of the computational structures as prepared for array programming with the support of optional parallel acceleration, is illustrated in \myfig{tag4}. To facilitate the processing of data within local areas, a user-customized tensor is proposed with an identical rank to that of the original data, to act as a generic container for an operator ($\boldsymbol{m}$ as depicted in \myfig{tag4}). In order to identify an integrated solution for the variable length of the ravel vector, a quasi-grid is designed as a computational component for calculating the necessary number of elements that should be ensured in the downstream calculation via the traversal on the tensor $\boldsymbol{x}$ under the action of operator $\boldsymbol{m}$. To illustrate, consider the following examples. In the context of global filtering, the requisite grid for elements is unquestionably the structure of the tensor $\boldsymbol{x}$ itself. But to the case of manipulations involving shrinkage, such as padding, the necessary grid will be the crossover points of orthogonal $k$-1 hyperplane families, which have been expanded with pre-defined stride distances along their coordinates. A gird tensor with new shape $\boldsymbol{s}'$ will be obtained following quasi-grid processing ($f_1$) on the shape of the tensor $\boldsymbol{x}$. The melt matrix $\boldsymbol{M}$ is a two-dimensional array with the number of rows in accordance with the prodcut of shape $\boldsymbol{s}'$, the number of columns equivalent to the length of the ravel vector of the operator $\boldsymbol{m}$. Each row of $\boldsymbol{M}$ is nothing else but a ravel vector determined by the $\boldsymbol{m}$-superpositioned region of tensor $\boldsymbol{x}$, as response to the movement of operator $\boldsymbol{m}$ on its grid mesh $f_1(\boldsymbol{x})$. In addition to the melt matrix $\boldsymbol{M}$, information such as the ravel vector $\boldsymbol{v}$ of operator $\boldsymbol{m}$ and the new shape $\boldsymbol{s}'$ of grid tensor is also included inside the intermediary structure. This is done for the facilitation for subsequent partition, broadcast operations in array programming, as well as further aggregation manipulations.\par

The profound significance for the design of $\boldsymbol{M}$ is that it is precisely the structure derived from a tensor that can not merely simultaneously satisfy the three constraints as listed in \mysec{Computational separability}, but also possesses the computational independence of its row vectors. A tensor could be flexibly subdivided under the precondition of ensuring partition rationality via melt matrix, which allows for the utilization of parallel acceleration further serving for mathematical manipulations, through the practical implementations analogous to that employed through multiple threadings, processes or distributed techniques.\par

It is counterintuitive that the partition of a tensor along any concrete dimension, instead of using its melt matrix, is invalid. This is because each numeric within a tensor is influenced by its neighborhood during its computational step. The operation of splitting a tensor along a coordinate dimension is certain to compromise the integrity of the associated neighbourhood relationship, which may potentially lead to methodological shortcomings in the generalization process to data with disparate dimensions. For example, utilising OpenCV, a toolkit designed for processing natural images (2-rank tensors), to process medical images is tantamount to conceding that tomographic images are all respectively independent. The information obtained from an OpenCV-filtered medical image is therefore prone to be merely the augmentation, on basis of its transversal projections instead of its real physical space.\par

\tagsubsection{Applied instances}

Design scheme of \mycode{tensor} sub-modular in our informatics project\mycite{chen2024informatics}, which aimed to provide a plethora of flexible meta implementations of mathematical computation on dense tensors, has proven to be a significant success of the programming framework in practice, particularly in the context of array programming applied to high-dimensional data. The present study selects and presents two exemplary algorithmic implementations that employ techniques such as high-dimensional generalization and array programming via melt matrix to concretize the essential concepts in the schematic diagram of the framework.\par

\subsubsection*{Bilateral filtering augmentation}

The bilateral filter was firstly proposed by C. Tomasi et. al.\mycite{tomasi1998bilateral} as a smoothing filter for image processing, with the feature of preserving edge-like areas.  As a non-linear method, the kernel of bilateral itself is a function of both the Gaussian distribution and the local elements in the tensor. Consequently, the computational complexity of the bilateral filter is greater than that of some other static kernels of the global filters. The calculation formula of this approach in the domain of natural image processing can be expressed as \myeq{tag5}:\par

\begin{equation}\label{tag5}
  \tilde{I}(x, y) = \frac{\sum_{j=-N}^{N} \sum_{i=-N}^{N} W(x, y, i, j) \cdot I(x+i, y+j)}{\sum_{j=-N}^{N} \sum_{i=-N}^{N} W(x, y, i, j)}
\end{equation}

Where the $x$ and $y$ denote absolute position of pixels of image to be filtered; $i$ and $j$ present the $(2N+1) \times (2N+1)$ neighbourhood areas, as relative to the center of a certain pixel $(x, y)$. $W(x, y, i, j)$ plays the role of the standard regulator, simultaneously as the function of both the pixel values and their associated neighbourhoods like:\par

\begin{equation}\label{tag6}
  W(x, y, i, j) = \exp \left(-\frac{(x-i)^2+(y-j)^2}{2\sigma_d^2}-\frac{\left\| I(x, y) - I(i, j) \right\|^2}{2\sigma_r^2}\right)
\end{equation}

In our study of the design proposals of bilateral filter-related APIs revealed that the majority of image processing libraries (e.g., OpenCV, scikit-image) impose limitations on input data (grayscale images or 2-dimensional images with channels) and oversimplify the anisotropy assumption of images to be filtered. In order to generalise all related concepts in \myeq{tag5} and (\ref{tag6}) into the context of tensor calculation, it is possible to utilise the position vector $\boldsymbol{x}$ of a tensor and its offset $\boldsymbol{s} = s(\boldsymbol{x})$ within neighbourhood to represent all the factors to be calculated. For example, the substitution of exponential items using $\boldsymbol{x}$ and $\boldsymbol{s}$ can be regarded as a generic implementation of the Bilateral filter designed in the Hilbert space for the processing on high-dimensional data:\par

\begin{equation}\label{eq3}
  W(\boldsymbol{x}, \boldsymbol{s}) \propto \exp \left( -\frac{(\boldsymbol{x}-\boldsymbol{s})^\top \boldsymbol{\Sigma}_d^{-1} (\boldsymbol{x}-\boldsymbol{s})}{2} - \frac{\left\| I(\boldsymbol{x}) - I(\boldsymbol{s}) \right\|^2}{2\sigma_r^2} \right)
\end{equation}

Proportion symbol $\propto$ in \myeq{eq3} herein is due to the normalization condition of weights (i.e., $W = W(\boldsymbol{x}, \boldsymbol{s})/\sum_{\boldsymbol{s}} W(\boldsymbol{x}, \boldsymbol{s})$), which must be satisfied prior to the broadcast on the melt matrix. $\boldsymbol{\Sigma}_d^{-1}$ and $\sigma_r^2$ represent types of matrices and numeric values, respectively, which represent the global and local deviations. The global $\boldsymbol{\Sigma}_d^{-1}$ can be used to provide the compatibility for anisotropy that should be considered in voxel-based computation.\par

\begin{figure}[htbp]
  \centering
  \includegraphics[scale=0.85]{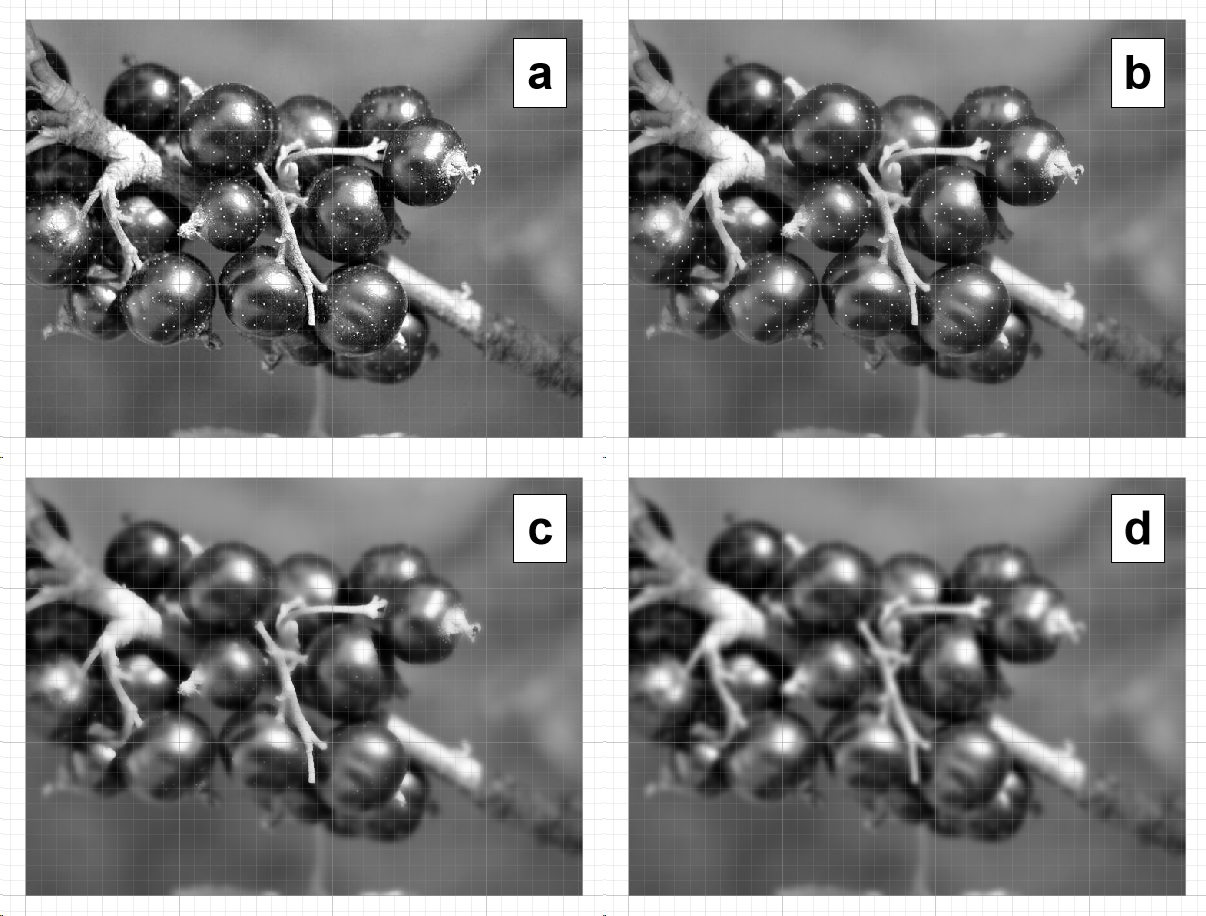}
  \caption{The (a) grayscale of a natural image, and its filtered results by bilateral filter using (b) local adaptive regulator $\sigma_r$; or constant regulator $\sigma_r$ with (c) appropriate and (d) excessive values respectively.}
  \label{tag7}
\end{figure}

In fact, there is a paucity of explicit definition regarding $\sigma_r$ with regard to its role as a local ranged regulator. In accordance with the motivation of \myeq{tag6}, the denominator $\sigma_d$ in the initial exponential item is designed to regulate the geometric shape of the neighbourhood operator, whereas $\sigma_r$ in the second one is for all local values $I(i,j)$. On basis of that consideration, the $\sigma_r$ should be a function of the grid point $\boldsymbol{x}$ (i.e., $\sigma_r = \sigma(\boldsymbol{x}, \boldsymbol{s})$), rather than the pre-defined argument as a constant. For the sake of simplicity, our implementation in informatics is designed to accommodate both alternatives. The following \myfig{tag7} illustrates the impact of filtering using different arguments of the bilateral filter on the example images.\par

The utilisation of adaptive $\sigma_r$ has an equivalent impact to that of applying a dynamic ruler to a scanned scope, which significantly removes local noise and makes effective purification of the original image, as demonstrated in \myfig{tag7} (b). As for the utilisation of constant $\sigma_r$ with scale on the approximate value to the norm of $\boldsymbol{\Sigma}_d$, the filtered result evinces a typical pattern of bilateral filtering result, which can be attributed to the combined effect of Gaussian de-noise and edge-perservation (as \myfig{tag7} (c) showed). \myfig{tag7} (d) is the case of a constant $\sigma_r$ with a value that is considerably greater than that of $\left\| \boldsymbol{\Sigma}_d \right\|^2$. The impact of the 2nd item in \myeq{eq3} becomes negligible, resulting in a final result that is similar to that of a Gaussian filter.\par

Aforementioned infrastructure shown in \myfig{tag4} offers a considerable degree of convenience in the fulfilment of a generic bilateral filter on the basis of \myeq{eq3} in an informatics project. Melt matrix of input tensor is generated by built-in function \mycode{\_pre\_generic\_map}, while the Gaussian component in \myeq{eq3} is determined by the \mycode{gaussian\_kernel} generator. For optional processing logic of local ranged deviation $\sigma_r$, despite the strategies of local adaptation or constant, the melt matrix can be broadcast into associated kernel matrices with the objective of calculating the subsequent results in the manner of handy array programming. This aspect of the work prospects in the contribution to the generic augmentation techniques, such as adaptive denoising, on high-dimensional data.\par

\subsubsection*{Gaussian curvature}

Curvature is a measurement that allows for the determination of the degree of variation.  Regions with high curvature values are often associated with high analytical values, such as in the case of valleys or peaks in tendency charts, or the corner-like areas in natural image processing. The Gaussian curvature, which is determined by the Hessian matrix in the context of natural images, can be presented as follows:\par

\begin{equation}\label{tag8}
  \boldsymbol{K} = \frac{\det [\boldsymbol{H}(x, y)]}{(1 + \boldsymbol{I}_x^2 + \boldsymbol{I}_y^2)^2}
\end{equation}

Where the $\boldsymbol{I} = \boldsymbol{I}(x, y)$ is for the input image. Subscript of $\boldsymbol{I}$ represents its partial differential in specific direction, e.g. $\boldsymbol{I}_{xy}$ is the 2nd-order differential $(\partial \boldsymbol{I})/(\partial x \partial y)$. Hessian is a matrix constituted by second-order differentials in all possible combinations of direction, as defined in \myeq{tag9}:\par

\begin{equation}\label{tag9}
  \boldsymbol{H}(x, y) = 
  \begin{bmatrix}
    \boldsymbol{I}_{xx}(x, y) & \boldsymbol{I}_{xy}(x, y) \\
    \boldsymbol{I}_{yx}(x, y) & \boldsymbol{I}_{yy}(x, y)
  \end{bmatrix}
\end{equation}

The symmetry of the Hessian matrix allows for the establishment of $\det (\boldsymbol{H}) = I_{xx} I_{yy} - I_{xy} I_{yx} = I_{xx} I_{yy} - I_{xy}^2$. If we extend the concepts above to a $m$-dimensional dense tensor $\textbf{I} \in \mathbb{R}^{d_1 \times \cdots \times d_m}$, the Gaussian curvature and the Hessian matrix will take the forms shown in \myeq{tag10} and (\ref{tag11}), respectively.\par

\begin{equation}\label{tag10}
  \textbf{K} = \frac{\det{[\textbf{H}(\textbf{I})]}}{(1+\sum_{i=1}^m \textbf{I}_{d_i}^2)^2}
\end{equation}

\begin{equation}\label{tag11}
  \textbf{H}(\textbf{I}) =
  \begin{bmatrix}
    \textbf{I}_{d_{1}d_{1}}  & \dots  & \textbf{I}_{d_{1}d_{m}} \\
    \vdots  & \ddots  & \vdots \\
    \textbf{I}_{d_{m}d_{1}} & \dots & \textbf{I}_{d_{m}d_{m}}
   \end{bmatrix}
\end{equation}

From the definition of $\textbf{H}(\textbf{I})$ in \myeq{tag11}, the data container used for computing Hessian is necessary to extend at least 2 dimension greater than $m$, due to the 2nd-order differentials. The complexity of its structure will easily result in obstacles on further computation. In contrast, the utilization of the melt matrix structure described in \mysec{Infrastructural architecture} enables the representation of a tensor with all neighborhood information that used for subsequent computations on gradients, in a much simplified format (2-rank tensor). This simple representation is also applicable for the first- and second-order partial differentials of that tensor. Thus, regardless of the actual dimension of a tensor in processing, all calculations can be reduced to occur within a tensor with ranks no greater than 4 (2 for melt matrix and the rest for saving first- and second-order differentials).\par

\begin{figure}[htbp]
  \centering
  \includegraphics[scale=0.85]{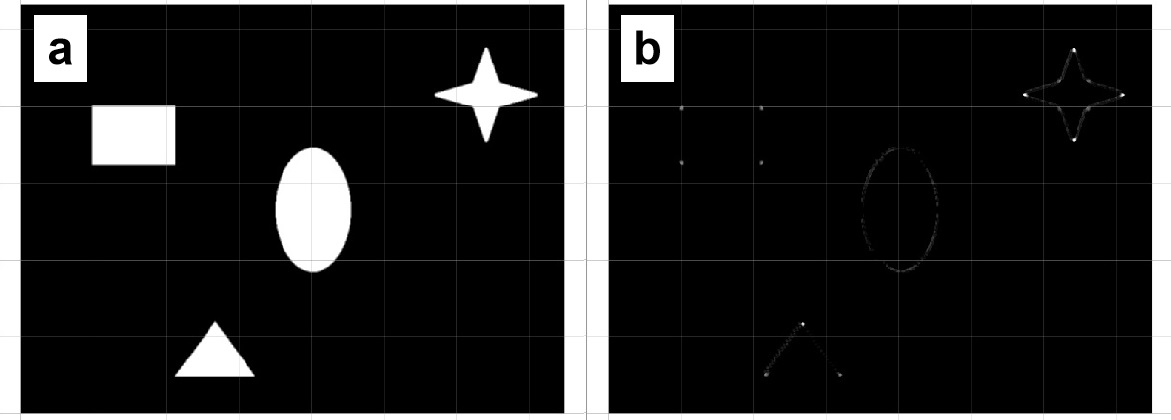}
  \caption{(a) 2-dimensional geometrical segmentation and (b) corresponding Gaussian curvature.}
  \label{tag12}
\end{figure}

\begin{figure}[htbp]
  \centering
  \includegraphics[scale=0.85]{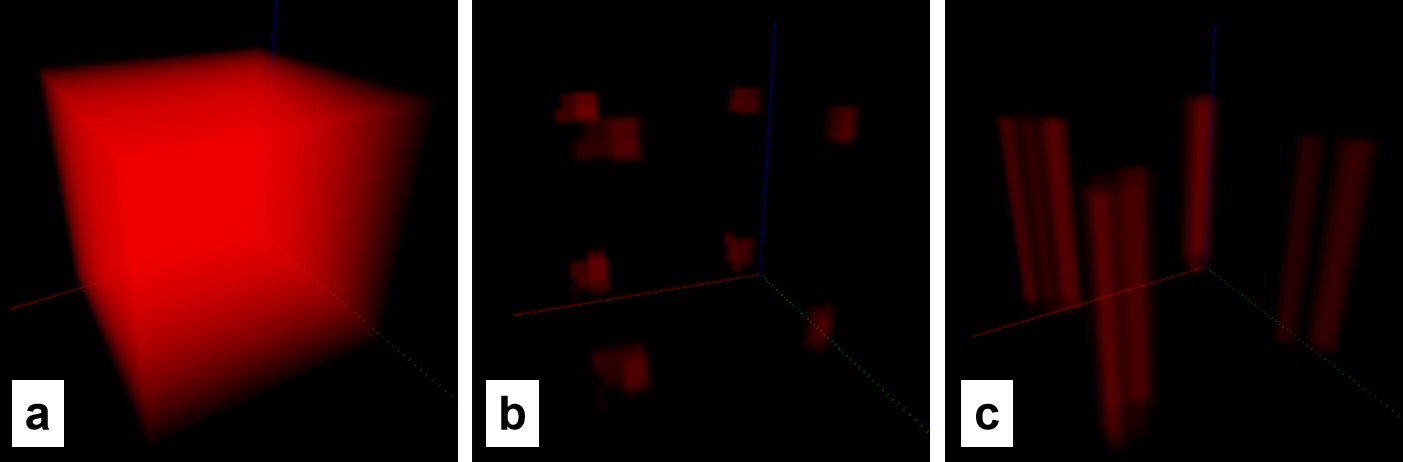}
  \caption{Monocolor renders of (a) 3-dimensional cube, and its (b) native 3-dimensional Gaussian curvature response, as well as (c) the 2-dimensional Gaussian responses stacked along $z$-axis.}
  \label{fig13}
\end{figure}

The design principles above permit the processing of tensors with varying ranks via an unified implementation. To illustrate, consider the two-dimensional segmentation depicted in \myfig{tag12} (a). The Gaussian curvature filter employed in the informatics project will markedly enhance all corner points, as evidenced in \myfig{tag12} (b). Similarly, the identical filter can also be applied to augment the vertices of a spatial cubic in three dimensions, as shown in \myfig{fig13} (a) and (b). For reference, we also demonstrate the result of forced application of planar operator on the tridimensional input, as illustrated in \myfig{fig13} (c).\par

Ignoring the dimension-related issues, taking oversimplification on high-dimensional data may possibly introduce tools that is inconsistent with the intrinsic dimension of the data to be manipulated. For example, expecting an augmentation for vertices of objects, but actually getting the augmentation of edges with a certain direction (e.g., the $z$-axis as illustrated in \myfig{fig13}). Through this case study, the demonstration here concretes the dimension-induced improper operations on data mentioned in \mysec{Infrastructural architecture}, which kind of issues should be worthy of comprehensive consideration by either the developers of architecture of scientific computation tools, or the users who apply that tools on their specific researches.\par

The removal of the dependence on data dimension can facilitate the generalization of the API in high-dimensional data contexts, and enable further computational optimizations (e.g., simplifying the computation of the $\textbf{H}(\textbf{I})$ via its symmetry, or partitioning on the melt matrix as prepared for using multi-node acceleration) in a more robust manner. This part of the work to compute the curvature response so far, is capable to contribute to techniques such as key point determination or spatial registration, on high-dimensional data.\par

\tagsection{Discussion}


Our research examined a potential solution for tasks of generic scientific computation on high-dimensional data, considering aspects such as theoretical completeness and technical implementability. Design paradigm of its infrastructural architecture suggested previously conforms to the mathematical abstraction of a tensor and simultaneously aligns with the single responsibility principle (SRP) of software engineering design. The mathematical abstraction and formulation for indefinite dimensional data structure enables generic functions to overcome the engineering challenges posed by dimensional mismatch across different applied domains. The proposal of the melt matrix and its related manipulations bring in not merely the computational facilitation through array broadcast, but moreover the computational reducibility on dimension into the underlying architecture of the system.\par

In spite of the generality in processing data with different structures, the execution time of a generic function will depend linearly on the number of elements that take part in computation. Different from the trade-off for introducing new observational dimensions into a data in analysis which should be considered by statisticians or domain-specific investigators, the phenomenon \emph{curse of dimensionality} indeed occurs as well, in the transition from mathematical representation to its concrete engineering implementation. With the melt matrix and its computational reducibility, the majority of the computational problems on high-dimensional data can be simplified with the constant time complexity. Whereas since overwhelming majority of computing operations occur in memory, for data in higher dimensions, the requisite space complexity is susceptible to exceeding the theoretical upper limit of a storage device with considerable probability. This limitation may originate from either the conceptual (e.g., threads and processes) or the physical (e.g., the performance of computing nodes) units for executing computation.\par

\begin{figure}[htbp]
  \centering
  \includegraphics[scale=0.5]{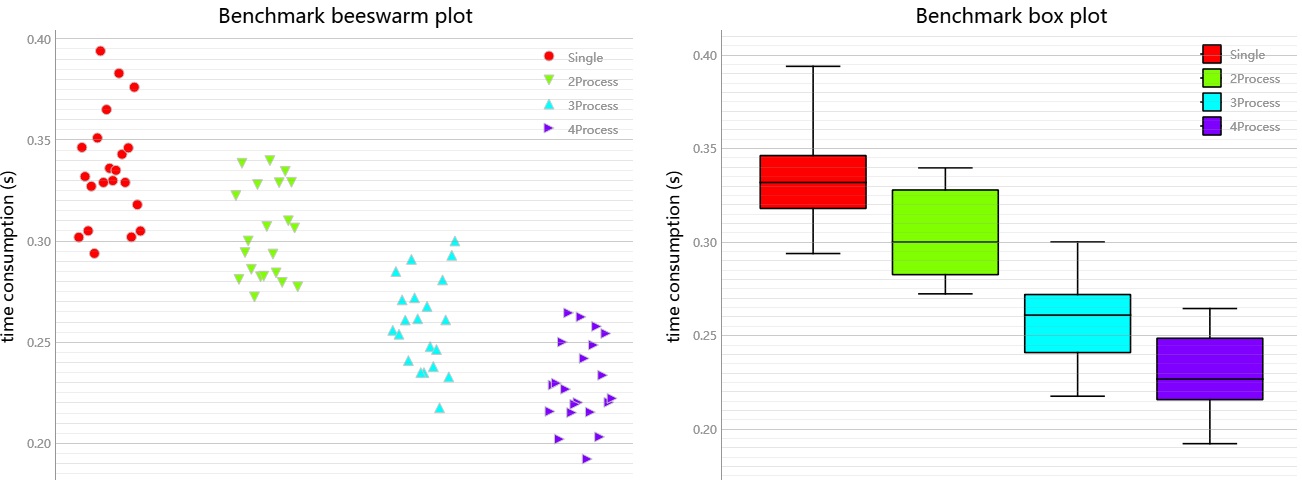}
  \caption{Beeswarm and box plots of benchmark test using single computing unit, 2, 3, and 4 processes as parallel units, labeled by \textbf{Single}, \textbf{2Process}, \textbf{3Process}, and \textbf{4Process} respectively.}
  \label{tag13}
\end{figure}

In order to ascertain the extent to which a parallel computing system might benefit from acceleration, we have devised a simple benchmark test of a global Gaussian filter applied to an identical 3-dimensional tensor. For purposes of comparison, the melt matrix derived from that tensor was also copied and partitioned into multiple matrix blocks in row-major. The practical time consumption during the execution of a multi-process parallelization is calculated by deducting the time spent in the process initialization and data partitioning from the total time cost. Trial in each experimental condition was subjected to 20 repetitions. It is inevitable to consider the impact of computing resource recovery when multiple processes are involved. In this context, the factual time cost associated with parallel acceleration will not be strictly inverse proportional to the number of processes involved. Nevertheless, the \myfig{tag13} illustrates a consistent decline in computing time with an increasing number of processes engaged in the computational process. Consequently, the utilization of parallel computation architecture is of at least two significant purposes: 1) to overcome the physical limitations of a single hardware system through the introduction of a computing cluster; and 2) to potentially improve the computational performance through parallel acceleration.\par

\begin{figure}[htbp]
  \centering
  \includegraphics[scale=0.5]{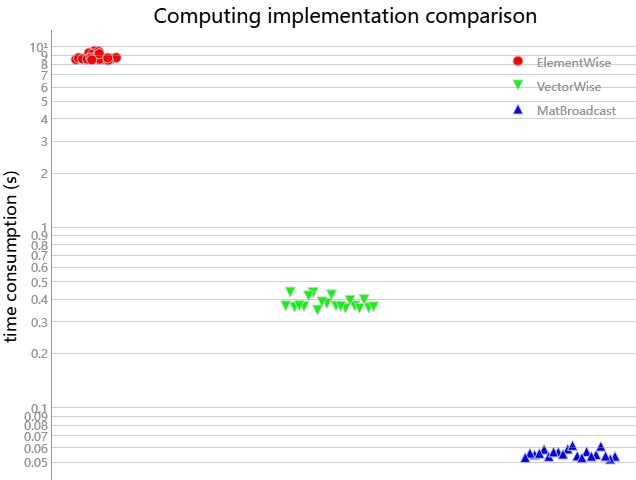}
  \caption{The time cost comparison of a Gaussian denoising process as for the computational implementations with different levels of abstraction: \textbf{ElementWise}, \textbf{VectorWise}, and \textbf{MatBroadcast} are the implementations using paradigms of respectively element-wise iteration, vector-wise iteration, and array programming broadcast on melt matrix.}
  \label{tag14}
\end{figure}

Moreover, due to the fact that implementations in different coding paradigms vary in running efficiency, a comprehensive understanding of performance with different implementations will be of significant importance in optimizing concrete computing steps. \myfig{tag14} illustrates the time required for the Gaussian kernel to be applied to the melt matrix in data processing, as depicted in \myfig{tag13}, using different paradigms of numerical iteration, vectorial iteration, and matrix broadcast. For the sake of convenient comparison, the logarithmic $y$-axis was employed. The degree of abstraction attained for the object undergoing iterative processing directly correlates with the efficiency of the computing implementation during execution. Specifically, the higher the abstraction level achieved, the more efficient the computational process becomes. In the instance where the matrix broadcast (designated as \textbf{MatBroadcast} in \myfig{tag14}) is chiefly employed, it has even been observed that such a method can yield execution speeds up to eight times faster than those achieved through the use of traditional vectorial iteration. In practical implementations, the significance of leveraging array programming therefore lies not merely in the facilitation of code expression on tensors through syntactic enhancements (also known as \emph{syntax sugar}), but more profoundly in the improvement on computational efficiency as provided by its native approaches.\par

In the exception of the previously mentioned array programming and melt matrix partitioning, another crucial factor in determining the potential performance of a computing system is the selection and management of underlying dependencies, in terms of infrastructural design. A well management of the dependencies will provide the connection to the current and future technologies. For example, the cupy project is developed in motivation of affording the completely coherent APIs of numpy and scipy to facilitate its running in GPU\mycite{harris2020array, virtanen2020scipy, nishino2017cupy}. Because of the rigourness in data exchange and iteroperability protocol of numpy and scipy teams, cupy developers can set the basic features of cupy array with considerably low cost, by merely overwriting two specific double-underscored methods (i.e. the \emph{dunder} method) of numpy array. Consistently, on the basis of powerful credit and technical strength of numpy, scipy and cupy teams, an increasingly perfect solution of computing system is under evolution with less technical obstacles.\par

\begin{figure}[htbp]
  \centering
  \includegraphics[scale=0.625]{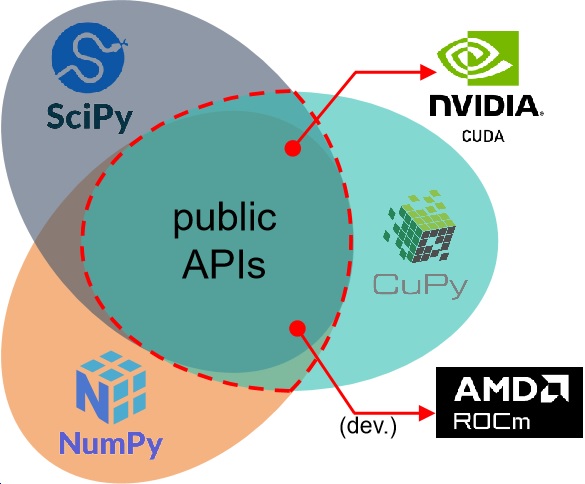}
  \caption{Venn diagram for the relationship of APIs provided by \mycode{numpy}, \mycode{scipy} and \mycode{cupy}.}
  \label{tag15}
\end{figure}

As a matter of fact, the majority of the implementations in the \mycode{tensor} module mentioned in \mysec{Applied instances}, including the aforementioned spatial bilateral filter and Gaussian curvature, are all in constructional design on the co-defined interfaces among numpy, scipy and cupy. This means the generic functions inside that module can even alter its computing backend as necessary (in another word, has the capability of further acceleration via NVIDIA CUDA or AMD ROCm). The design in programming paradigm based on the public APIs of those three packages, precisely the $\boldsymbol{S}_{\mathrm{cupy}} \cap (\boldsymbol{S}_{\mathrm{numpy}} \cup \boldsymbol{S}_{\mathrm{scipy}})$ as illustrated in \myfig{tag15}, is generally in deserving of preference, through which implementation a computing system can own the forwards-compatibility with the continuously optimized underlying libraries, as well as with the ecology in increasing maturity as built by GPU suppliers.\par

\tagsection{Conclusion}

This study is specializing in the generic solution right now and forwards, for mathematical computations on high-dimensional data. By examinig the mathematical properties that should be incorporated into an infrastructural component for general computation, we have conceptualized an intermediate structure (melt matrix) of the generic container (tensor).  This design paradigm satisfies both the Hilbert completeness and computational separability criteria, thereby providing theoretical and practical support for generic function construction and the feasibility of applying parallel acceleration techniques.\par

To concrete the practical advantages of the coding implementation based on the proposed paradigm, the paper also presents two typical applications used primarily in the field of computer vision. Through the methodological universalization of their respective mathematical principles allows for the fulfillment of both practical coding implementations with an unified infrastructure composed essentially of melt matrix and associated calculations. As for the sight of computational reducibility, the melt matrix can equivalently transform the computation on dense tensor with any shape into that of a 2-rank tensor, which establishes the implementary invariance as uncorrelated to dimensionality. The generic functions obeyed this paradigm are in a sense the general solutions with forward-compatibility design for their respective domains. And the computing system as built above those generic functions is consequently low risky in accumulating technical debt with the progress of its own iteration and evolution. Besides, another important feature of the melt matrix is computational independence of the row items which make the set partitioning in mathematics' context be capable on it. In practical engineering, such kind of underlying design provides the feasibility as well as considerable convenience for split and parallel processing of a computing task. It technically bridges the pathway meanwhile, from scientific computations on tensors to the approaches of modern parallel acceleration.\par

In addition to the further evaluations of the performance improvements achieved by parallel acceleration and array programming paradigm in the discussion, an investigation into the Python-major GPU computing ecology has also been conducted by us. Drawing upon the design principles grounded in rigorous scientific arguments, we have distilled an avant-garde design paradigm from our ongoing project and coding practice, as outlined in this paper.             This paradigm not only produces persistent contribution on our engineering realizations, but also provides a future-oriented architecture design of software for solving high-dimensional problems.\par

\subsection*{Acknowledgement}

The authors would like to express their sincere gratitude to the support from Central Research Institute (CRI) at United Imaging Healthcare (UIH) group.    Special thanks are due to the Precision Medicine Joint Laboratory (PMJL) at Ruijin hospital United imaging Joint Center of research and development (RUJC) for part of the requisite resources. The authors also acknowledge the constructive comments of their colleagues whose efforts were invaluable during the course of this study.\par

Additionally, the authors would like to extend our highest consideration and respect to the Python community, the teams of NumPy, SciPy, and CuPy as well, for their collective efforts in creating and maintaining these invaluable tools have significantly contributed to the success of our research. Their dedication to open-source software and scientific computing has enabled us to conduct our work efficiently and effectively. We are deeply appreciative of their contributions and the impact they have had on the research community.\par

\subsection*{Data availability}

The underlying data is available on \href{https://pixnio.com/}{pixnio.com} from the public domain, under the terms of the \href{https://creativecommons.org/public-domain/cc0/}{Creative Commons 0}.\par

\subsection*{Code availability}

All code for implementations of framework components and generic functions associated with the current submission is available in the \href{https://informatics.readthedocs.io/en/latest/interface/api\_misc.html\#misc-modules-and-functions}{\emph{kernel utilities}} module and the \href{https://informatics.readthedocs.io/en/latest/interface/api\_tensor.html\#module-tensor}{\emph{tensor}} subpackage of the \href{https://pypi.org/project/informatics/}{\emph{informatics}} project (currently version 0.0.5rc1), released under the terms of the \href{https://www.apache.org/licenses/LICENSE-2.0.html}{Apache Software License v2.0}.\par

\clearpage
\bibliographystyle{unsrt}
\bibliography{full}
\end{document}